\documentclass[runningheads]{llncs}
\usepackage[T1]{fontenc}
\usepackage{graphicx}
\usepackage{amsmath}
\usepackage{hyperref}

\newcommand{\orcid}[1]{\href{https://orcid.org/#1}{\includegraphics[width=0.03\textwidth]{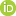}}}

\begin{document}
%
\title{FPGA-Based Hardware Architecture for Contrast Maximization in Event-Based Vision}
\titlerunning{Contrast Maximization hardware architecture}
%
%
\author{Michał Filipkowski \and
Marcin Kowalczyk\orcid{1111-2222-3333-4444} and
Tomasz Kryjak\orcid{2222--3333-4444-5555}}
%
\authorrunning{M. Filipkowski et al.}
%
\institute{AGH University of Krakow, Poland\\ Embedded Vision Systems Group, Computer Vision Laboratory
\email{mfilipkowski@student.agh.edu.pl \\
\{kowalczyk, kryjak\}@agh.edu.pl} }
\maketitle              
\begin{abstract}
This paper presents a~hardware architecture that implements the Contrast Maximization (CM) algorithm in Field-Programmable Gate Array (FPGA) resources for event-based vision systems. CM estimates motion parameters by maximizing the contrast of an Image of Warped Events (IWE) reconstructed from asynchronous event streams. Event-based vision sensors generate sparse data with high temporal resolution and low spatial redundancy, which makes them well suited for hardware processing.
The deterministic, massively parallel structure of the FPGA is leveraged to design a deeply pipelined architecture capable of high-throughput, energy-efficient processing suitable for real-time embedded applications. This paper details the hardware modules responsible for event warping, contrast computation, and iterative optimization, discusses key implementation decisions, and presents the hardware-aware optimization method used in the design.
Experimental results demonstrate a substantial speed and efficiency improvement over CPU- and GPU-based implementations, with motion parameter estimation executing over 200 times faster. To the best of our knowledge, this is the first hardware architecture enabling acceleration of CM algorithm computations.
Its performance is evaluated in terms of processing speed, energy efficiency, and hardware resource utilization.
The proposed design is validated using an event-based object tracking application.
The results confirm that the architecture provides a solid foundation for real-time motion estimation in high-speed, low-power embedded systems.

\keywords{FPGA \and Contrast Maximization \and \mbox{Event-based camera} \and Object tracking \and Neuromorphic vision}
\end{abstract}
\section{Introduction}

The rapid development of modern sensing technologies has made event-based cameras, also known as dynamic vision sensors (DVS), an increasingly popular research topic in computer vision and robotics. Event cameras are bio-inspired sensors with many advantages over frame-based cameras. The superior features include independent (asynchronous) pixel operation, high temporal resolution, high dynamic range, relatively small motion blur, and latency on the order of microseconds~\cite{gallego2022survey,gehrig2018asynchronous}. Event cameras operate by detecting changes in pixel brightness rather than continuously capturing the full image at fixed intervals, like a typical frame-based camera.
This sparse nature makes them particularly well suited for real-time applications with high-speed motion, such as Advanced Driver Assistance Systems, Autonomous Vehicles, and drone navigation~\cite{gallego2022survey,maqueda2018steering}.

However, processing data from event cameras usually requires specialized algorithms that can handle high temporal resolution and irregular input patterns or converting the data to a dense representation.
One of the techniques used to process such data is the Contrast Maximization (CM) algorithm, which estimates motion parameters.
This method may serve as the foundation for many tasks, such as visual odometry, optical flow, and depth estimation in event-based vision \cite{gallego2018unifying,guo2024cmax}.
However, CM requires the use of an iterative optimization algorithm to maximize the objective function that describes the sharpness of a warped event frame. As a result, this method is characterized by high computational complexity, which translates into high power consumption and difficulty in real-time operation, even on PC-class computers.
For this reason, we decided to design a~dedicated hardware architecture and implement it using Field Programmable Gate Array (FPGA) resources.

The high performance of FPGAs stems from their flexibility and fine grained parallelism, enabling us to create a fully optimized circuit for a given application ~\cite{asano2009comparison}. It means that complex operations such as warping events, calculating Image of Warped Events (IWE) and gradients could be executed simultaneously with the architecture adapted to achieve the best performance for a given algorithm.
For complete vision pipelines, FPGAs often outperform Graphics Processing Units (GPUs) and Central Processing Units (CPUs), and their advantage increases as the complexity of the vision pipeline grows~\cite{qasaimeh2019comparing}. However, there are some constraints in developing applications on FPGAs. First, the design process usually requires more effort than implementing the same algorithm on a~CPU and GPU.
Second, internal memory resources in FPGAs are rather limited, which means that it is not possible to store large amounts of data in memory and requires careful design of the architecture.

In this work, an object tracking application is selected to verify how the CM algorithm works under practical conditions. The model estimates the translation of the target and fits the motion parameters, which in turn causes the region of interest (ROI) to be updated. This means that the application evaluates the horizontal and vertical velocity components of the tracked object, enabling accurate motion estimation in event-based data.

The main contribution of this paper is the design of an architecture realizing the contrast maximization algorithm on an FPGA. The design includes event preprocessing, event warping, calculating gradients, running an optimization loop that finds the proper motion parameters, and updating ROI with the result of the CM method.
The achieved parameters are compared with the CPU and GPU implementations realizing the same CM algorithm in order to provide an evaluation of the results achieved. The proposed architecture allowed for processing the data over 200 times faster than the CPU or GPU implementations while consuming under 1 W of on-chip power.

The remainder of this paper is organized as follows. In Section \ref{sec:related_work} related works are presented. Section \ref{sec:methodology} contains the methodology of a CM algorithm and its mathematical model. Section \ref{sec:architecture} describes the proposed FPGA architecture. In Section \ref{sec:results} the experimental results are presented. Finally, section \ref{sec:conclusions} presents the conclusions with a discussion of future research directions.

\section{Related Work}
\label{sec:related_work}
The CM algorithm is a popular approach in event data processing. It enables the estimation of motion, depth and optical flow by warping events and maximizing IWE variance~\cite{gallego2018unifying}. Furthermore, CM is employed in event-based visual odometry to enable real-time estimation of camera trajectories, for example in methods based on IWE and geometric optimization with contrast maximization in the volumetric ray field~\cite{wang2022visual}. A review of the most commonly used cost functions in the context of CM-based implementations has also been conducted~\cite{stoffregen2019event}. In addition, several modifications to the classical CM framework have been proposed to improve its performance, including multi-scale warping to enhance convergence and avoid local minima, as well as geometric regularizers to prevent event collapse~\cite{shiba2022event,shiba2024secrets}. Most work to date uses CM implementations running on CPUs, which typically do not operate in full real- time~\cite{liu2020globally,peng2021globally}.
However, there are also examples of the algorithm being implemented in real-time on a~CPU. Using globally aligned event data reduces drift and improves IWE contrast in longer time windows~\cite{kim2021real}. Despite the relatively high computational complexity, stable system performance can be maintained through the use of properly organized event buffers and optimized methods of combining them.


The literature also describes solutions that implement CM using the Compute Unified Device Architecture (CUDA) acceleration. Integration of CM with sequential learning methods and optical flow estimation techniques is also becoming increasingly common. Running these algorithms on GPUs enables very high inference speeds and operation in near real-time conditions~\cite{hamann2024motion,paredes2023taming}.

Our review of the scientific literature revealed no examples of architectures implementing the CM algorithm on an FPGA platform. This allows us to conclude that the presented solution is the first of its kind. The results confirm the real possibility of accelerating event processing using the CM algorithm, and demonstrate that the approach presented in this article effectively achieves this goal.

\section{Methodology}
\label{sec:methodology}
This section presents the methodology used in the CM algorithm. 

\subsection{Events representation and warping}
\label{ssec:representation}
Events cameras have independent pixels that operate continuously and generate events when a brightness change is detected. Each event $e_k = (t_k, x_k, y_k, p_k)$ contains the timestamp when the event occurred, the coordinates $(x_k, y_k)$ and the polarity of the change $p_k \in \{-1, 1\}$. In most event datasets, the polarity has a value $p_k = 0$, when the brightness of the pixel decreases.
Events occur with a variable rate that depends on the scene dynamics.
In the CM algorithm, events are processed in successive batches and warped to a selected reference time based on the motion model, so that edges and textures in the observed scene become maximally aligned. Several strategies for choosing this reference time exist. In our experiments we selected a simple model, where the reference time is defined as the midpoint of the batch. This is described in the Eq. \eqref{eq:tref}.

\begin{equation}
    t_{ref} = t_1 + (t_{N_e}-t_1)/2
    \label{eq:tref}
\end{equation}
where: \(t_1\) and \(t_{N_e}\) are the minimum and maximum times in the event batch, respectively. $N_e$ denotes the number of events in a given batch.

Each event in the batch is then moved to the reference time according to the adopted motion model and its parameters. The spatial coordinates of the events are then propagated forward or backward in time, based on the predicted displacement according to the model. This study uses a two-dimensional motion model with a constant translation speed along the x and y axes. To this end, the difference between the reference time ($t_{ref}$) and the timestamp ($t_k$) is determined for each event, as described by Equation \eqref{eq:dt}.

\begin{equation}
    dt = t_k - t_{ref}
    \label{eq:dt}
\end{equation}

Then, the new position of each event is determined according to Eq. \eqref{eq:e'_k} and \eqref{eq:x'k}, based on the timestamp difference and motion parameters \((v_x, v_y)\).
\begin{equation}
    e_k = (t_k, x_k, y_k, p_k) \:\rightarrow\: e'_k = (dt_k, x'_k, y'_k, p_k)
    \label{eq:e'_k}
\end{equation}
\begin{equation}
    x'_k = x_k - dt_k \cdot v_x
    \label{eq:x'k}
\qquad
    y'_k = y_k - dt_k \cdot v_y
\end{equation}

The warping step compensates for edge movement in the scene, ensuring that events originating from the same physical edge converge at consistent spatial locations at time \(t_{ref}\). The quality of the result is directly dependent on the accuracy of the motion model used.

\subsection{Image of Warped Events}
\label{ssec:iwe}

Once all events have been transformed to the reference time, the next step in the algorithm is to create a warped event frame. This involves superimposing each transformed event on a two-dimensional pixel grid to create an image representing the density of events after motion compensation.

\begin{figure}[htbp]
    \centering
    \includegraphics[width=0.35\textwidth,keepaspectratio]{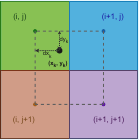}
    \caption{Visualization of bilinear voting operation}
    \label{fig1}
\end{figure}

After warping events, their spatial positions often fall between discrete pixel locations. Arbitrary assignment of events to the nearest pixel can lead to significant inaccuracies when constructing the IWE. Therefore, a bilinear voting approach is used in which each of the four neighboring pixels receives a contribution proportional to its distance from the warped event. In practice, this is achieved by extracting the fractional parts of the event coordinates $\forall(x_k, y_k)$ and using them to compute the weights that determine how the event value is distributed among the surrounding pixels, as in Eq. \eqref{eq:weights}.

\begin{equation}
w_{k,i,j} =
\begin{cases}
(1 - dx_k)\cdot (1 - dy_k) & \text{for pixel } (i_k,j_k) \\
dx_k \cdot (1 - dy_k) & \text{for pixel } (i_k+1,j_k) \\
(1 - dx_k)\cdot dy_k & \text{for pixel } (i_k,j_k+1) \\
dx_k \cdot dy_k & \text{for pixel } (i_k+1,j_k+1)
\end{cases}
\label{eq:weights}
\end{equation}
where: $w_{k,i,j}$ denotes the weights assigned to the neighboring pixels of the event \(e'_k\), and $dx_k$, $dy_k$ represent the fractional distances from the integer pixel coordinate $(i_k, j_k)$ along the $x$ and $y$ axes: \(dx_k = x'_k - i_k\), \(dy_k = y'_k - j_k\).
This operation is visualized in Fig. \ref{fig1}.

Summing the coefficients \(w_{k,i,j}\) for all pixels gives us the image of warped events (IWE), described by Equation \eqref{eq:I_w}.

\begin{equation}
    I_{W}(i, j) = \sum_{k=1}^{N_e} w_{k,i,j}
    \label{eq:I_w}
\end{equation}

\subsection{Objective function}
\label{ssec:objectivefunc}
The resulting image is used to calculate the contrast function, which is then used as the objective function in the optimization process. Contrast can be defined in various ways. These definitions are described and compared in \cite{gallego2019focus}. A commonly used measure is the variance of pixel values in the generated image. Intuitively, the better the motion parameters are chosen, the more events associated with the same physical edge will overlap, resulting in a clearer, sharper image with increased variance. The objective function is described by Eq. \eqref{eq:C}.

\begin{equation}
    C = Var(I_{w}) = \frac{1}{N_p} \sum_{(i, j) \in \Omega} (I_w(i, j) - \mu)^2 \qquad
    \mu = \frac{1}{N_p} \sum_{(i, j) \in \Omega} I_w(i, j)
    \label{eq:C}
\end{equation}
where \(N_p\) is the number of pixels and \(\Omega\) is the set of all pixels, so \(N_p = | \Omega |\).


Hence, the value of the contrast function reflects the accuracy of the current motion model: the higher the value, the better the parameters \(v_x\) and \(v_y\) compensate for the actual edge displacement in the scene. The optimization algorithm uses this value to update the motion parameters, iteratively seeking their optimal combination.

\subsection{Gradient calculation}
\label{ssec:gradient}
To maximize the contrast function effectively, it is best to use a gradient optimization method, which uses the derivatives of the objective function to determine the direction of the fastest increase in contrast. To do this, it is necessary to determine the gradient of the contrast function with respect to the motion parameters, which is given by Eq. \eqref{eq:gradient}.

\begin{equation}
    \nabla C = \left[ \frac{\partial C}{\partial v_x}, \frac{\partial C}{\partial v_y} \right]^T
    \label{eq:gradient}
\end{equation}


Changes in the position of events affect the pixel values of the warped image and, consequently, the contrast function. The gradient can be calculated using the chain rule, as described by Eq. \eqref{eq:chainC} for the \(x\)-axis.

\begin{equation}
    \begin{split}
        \frac{\partial C}{\partial v_x} &= \frac{\partial C}{\partial I_w} \cdot \frac{\partial I_w}{\partial x'} \cdot \frac{\partial x'}{\partial v_x} \\
    \end{split}
    \label{eq:chainC}
\end{equation}

Then we obtain Eq. \eqref{eq:C_v_x}.
\begin{equation}
    \frac{\partial C}{\partial v_x} = \frac{2}{N_p} \sum_{(i, j) \in \Omega}(I_w(i,j; v_x) - \mu(v_x)) (\frac{\partial I_w(i, j; v_x)}{\partial v_x} - \frac{\mu (v_x)}{\partial v_x}
    \label{eq:C_v_x})
\end{equation}

The derivative of the mean value \(I_w\) can be expressed according to Eq. \eqref{eq:mu_v_x}.
\begin{equation}
    \frac{\partial \mu(v_x)}{\partial v_x} = \frac{1}{N_p} \sum_{(i,j) \in \Omega} \frac{\partial I_w(i, j; v_x)}{\partial v_x}
    \label{eq:mu_v_x}
\end{equation}

This is therefore the average value of the derivative \(I_w\) with respect to velocity in a given direction. We use Equation \eqref{eq:I_w} to calculate the derivative \(I_w\) with respect to velocity, obtaining Equation \eqref{eq:I_w_v_x}.
\begin{equation}
    \frac{\partial I_w(i, j)}{\partial v_x} = \sum_{k=1}^{N_e}\frac{\partial w_{k, i, j}}{\partial dx_k} \cdot \frac{\partial dx_k}{\partial x'_k} \cdot \frac{\partial x'_k}{\partial v_x}
    \label{eq:I_w_v_x}
\end{equation}

Based on the previous formulas, we can calculate the next necessary elements, obtaining Eq. \eqref{eq:w_dx_x} and \eqref{eq:dx_k_x'_k_x'_k_v_x}.

\begin{equation}
\frac{\partial w_{k, i, j}}{\partial dx_k} =
\begin{cases}
-(1 - dy_k) & \text{pixel } (i,j) \\
(1 - dy_k) & \text{pixel } (i+1,j) \\
-dy_k & \text{pixel } (i,j+1) \\
dy_k & \text{pixel } (i+1,j+1)
\end{cases}
\label{eq:w_dx_x}
\end{equation}

\begin{equation}
    \frac{\partial dx_k}{\partial x'_k} = 1
\qquad
    \frac{\partial x'_k}{\partial v_x} = -dt_k
    \label{eq:dx_k_x'_k_x'_k_v_x}
\end{equation}

Corresponding patterns can also be determined for the \(y\)-axis. From the above relations, it follows that calculating the gradient requires three images to be determined: IWE from Eq. \eqref{eq:I_w} and two derivative images for \(v_x\)  and \(v_y\) according to Eq. \eqref{eq:I_w_v_x}. These images are obtained by summing the values corresponding to successive processed events. Then, according to Eq. \eqref{eq:C_v_x}, we subtract the average values of these images, multiply them together and sum them up.




\subsection{Optimization loop}
\label{ssec:optimization}


With the gradient of the objective function determined, the gradient ascent method was chosen to find the motion parameters that maximize the objective function.
For this purpose, in each iteration of the optimization loop, a new approximation of the solution \(v_x\) and \(v_y\) is designated. This is represented by Eq. \eqref{eq:v_update}.

\begin{equation}
\begin{bmatrix} v_x^{(n+1)} \\ v_y^{(n+1)} \end{bmatrix} 
=
\begin{bmatrix} v_x^{(n)} \\ v_y^{(n)} \end{bmatrix} 
+ \eta \cdot \nabla C^{(n)}
\label{eq:v_update}
\end{equation} 
where: \(\eta\) denotes the learning rate of the gradient ascent algorithm, and \(n\) is the number of iterations of the optimization loop.


When the maximum number of iterations \(T\) is reached, the coordinates of the ROI \((x_{roi}, y_{roi})\) are updated based on the calculated motion parameters in the example object tracking application. This is shown in Eq. \eqref{eq:roi_update}. Then, processing of the next batch of events begins.

\begin{equation}
\begin{bmatrix} x_{roi}^{(N_b+1)} \\ y_{roi}^{(N_b+1)} \end{bmatrix} 
=
\begin{bmatrix} x_{roi}^{(N_b)} \\ y_{roi}^{(N_b)} \end{bmatrix} 
+ 
\begin{bmatrix} v_x^{(T)} \\ v_y^{(T)} \end{bmatrix} 
\label{eq:roi_update}
\end{equation}
where: \(N_b\) denotes batch number.

\begin{figure}[h!]
    \centering
    \includegraphics[width=1\textwidth,keepaspectratio]{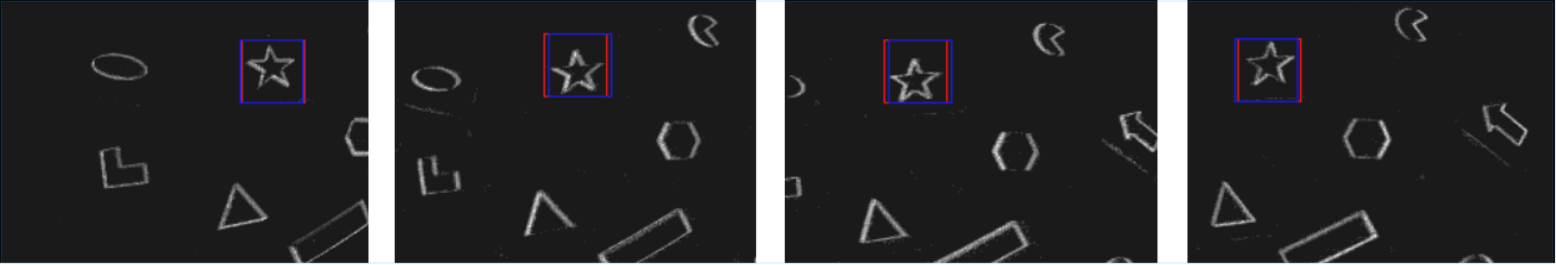}
    \caption{Sequence of event frames from DAVIS 240C datasets~\cite{mueggler2017event} demonstrating the CM-based tracking and ROI update process.}
    \label{fig2}
\end{figure}

Fig. \ref{fig2} shows a visualization of the ROI update process between specific event batches. Accumulated events are represented by white dots, while the red frame indicates the current position of the ROI window.

\section{FPGA Architecture Design}
\label{sec:architecture}

\begin{figure}[h!]
    \centering
    \includegraphics[width=0.73\textwidth,keepaspectratio]{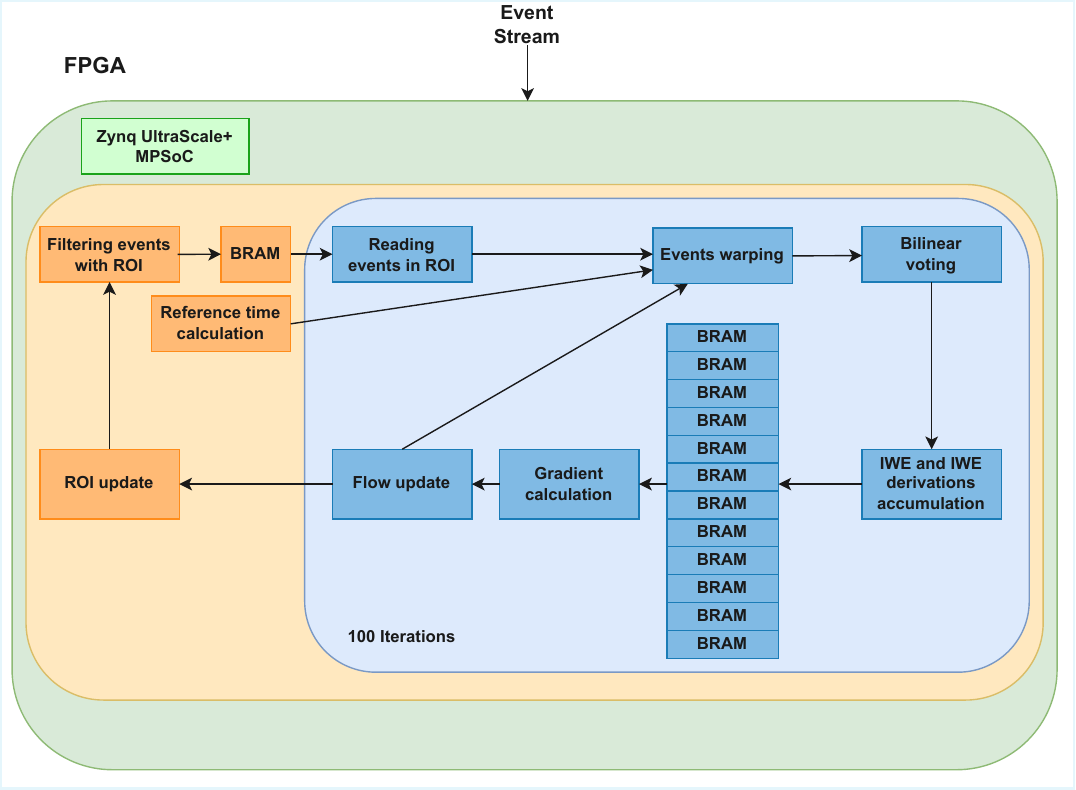}
    \caption{Hardware architecture diagram of an application for object tracking using CM}
    \label{fig3}
\end{figure}

\begin{figure}[htbp]
    \centering
    \includegraphics[width=0.4\textwidth,keepaspectratio]{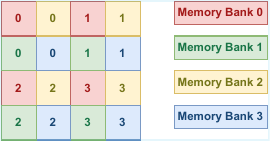}
    \caption{Visualization of memory management for bilinear voting weights}
    \label{fig4}
\end{figure}

We designed hardware architecture to realize the CM algorithm for an example application of event-based object tracking. Its diagram is shown in Fig. \ref{fig3}. The first stage of the pipeline is the preprocessing of input events. Each event is checked to determine whether it lies within the ROI. Events that satisfy this condition are stored in BRAM. In parallel with the event selection, the reference time is calculated according to Eq. \eqref{eq:tref} and the range of the timestamp differences is also scaled to \([-1, 1]\).
Once the preprocessing stage is complete, events are read back from BRAM and passed to the warping module. This begins the stream-processing phase, where events from the current batch are processed over 100 optimization iterations to estimate the motion. The number of iterations was selected based on the experiments conducted. It was enough for the optimization algorithms to converge within this number of iterations, but it could be easily modified in the architecture. Warped events, computed according to Equations (\ref{eq:dt} - \ref{eq:x'k}), are forwarded to the bilinear voting module.
In the voting stage, each warped event is distributed across 4 neighboring pixels using bilinear weights. The resulting contributions are written in 12 separate BRAM instances: four store the values of \(I_w\), four accumulate the partial derivatives with respect to \(v_x\), and the remaining four accumulate the derivatives with respect to \(v_y\). Each group of 4 memory banks corresponds to a specific subset of ROI pixels, partitioned according to the parity of their coordinates. The parity information, together with the appropriate target addresses, is determined relative to the starting point of the ROI. This division of memory into 12 banks enables simultaneous writing of all contributions within a single clock cycle.
The complete memory management and addressing scheme is illustrated in Fig. \ref{fig4}.

The accumulation of pixel values is performed in three pipeline stages. In the first cycle, the target address for the incoming weight is provided to the module, and the current value stored at this memory location is read from BRAM. In the second cycle, this previously read value is added to the new weight entering the module. In the third cycle, the resulting accumulated value is written back to the appropriate BRAM cell.
This pipeline introduces the risk of write conflicts: if two events refer to the same memory address within a few consecutive clock cycles, the first event may be overwritten. This occurs because its updated value may not be written to BRAM before the next event triggers a read of the same address.
To avoid this issue, a small register-based buffer was added to store the addresses and accumulated values of the three most recently processed entries in the accumulation module. If the address of a new event matches any of the buffered addresses, the value is retrieved from the corresponding register instead of from BRAM. The new weight is then added to the buffered value, ensuring correctness before the result is written back.
For both image derivatives, the accumulation mechanism operates in exactly the same manner.

Once the accumulation of pixel values in all memory blocks is complete, the gradient is computed according to Eq. \eqref{eq:C_v_x}. An important implementation detail is that a true hardware reset of the BRAM is not possible. Therefore, one cycle after reading the value required for gradient evaluation at a given address, a zero is written back to this location. This ensures that before the current algorithm iteration finishes, all BRAM instances are already cleared and ready to accumulate new bilinear weights in the next iteration.
The optimization iteration concludes with the update of the horizontal and vertical motion components, as defined in Eq. \eqref{eq:v_update}. The next optimization loop then begins, in which the events are warped again using the newly updated velocities.

When the maximum number of iterations is reached, the computational module outputs the flow values obtained in the final iteration. These motion estimates are subsequently used to update the ROI according to Eq. \eqref{eq:roi_update}.

\section{Experimental results}
\label{sec:results}
The designed architecture was tested on the \textit{Kria KV260 Vision AI Starter Kit} and verified for consistency against a Python reference model developed in PyTorch. The computation results were inspected using the \textit{Integrated Logic Analyzer (ILA)}. Based on an analysis of the architecture, Eq. \eqref{eq:C_clk} was derived to describe the number of clock cycles required to process a single batch of events.

\begin{equation}
C_{batch}(N, T, n, P) = N + T \cdot(n + L_r + \frac{P}{4} + L_v)
\label{eq:C_clk}
\end{equation}
where: \(N\) is the total number of events in a batch, \(T\) is the number of optimization iterations, \(n\) is the number of events within the ROI, \(L_r\) is the latency associated with processing readouts from the ROI (32 cycles), \(P\) is the total number of ROI pixels, and \(L_v\) is the latency of the bilinear voting operations (35 cycles).

Therefore, the total processing time for a single batch can be calculated as \(C_{batch}/f_{clk}\), where \(f_{clk}\) is the clock frequency of the architecture.
The system achieved a maximum clock frequency of 210 MHz.
A comparison was then performed for the time required to process a batch of 5,000 events, assuming an ROI resolution of \(64 \times 64\) pixels, 800 events in the ROI (for the batch used in the comparison), and 100 iterations of the optimization loop. The reference model was executed on an \textit{Intel Core i5-11300H 3.10 GHz CPU} and an \textit{Nvidia GeForce RTX 3050 Ti GPU}.
The execution time of the proposed FPGA architecture was 0.92 ms, compared to 185.96 ms on the CPU and 473.51 ms on the GPU.
This improvement stems from leveraging parallel computation and efficient memory management strategies within the FPGA architecture.

These results can also be compared with the computation times reported in \cite{kim2021real}. The authors provide processing times of approximately 30–60 ms for 90 optimization iterations, 5000 events, and a resolution of \(240 \times 180\) on an \textit{i7-7500U@2.7GHz} CPU. For the same parameters, the processing time of the proposed architecture operating at a 200 MHz clock can be estimated using Eq.~\eqref{eq:C_clk} as 7.2 ms. It should be noted, however, that supporting this resolution with the presented architecture would require an FPGA device equipped with more BRAM resources than the one used in this work.


\begin{table}[h!]
\centering
\caption{Resource utilization summary for the CM algorithm on FPGA}
\label{tab:main_module_resources}
    \begin{tabular}{|l|r|}
    \hline
    \textbf{Resource} & \textbf{Count} \\
    \hline
    CLB LUTs & 3655 (3.12\%) \\
    CLB Registers & 7857 (3.35\%) \\
    CLB & 1218 (8.32\%)\\
    \hline
    \end{tabular}%
    \hspace{1cm} 
    \begin{tabular}{|l|r|}
    \hline
    \textbf{Resource} & \textbf{Count} \\
    \hline
    Block RAM & 25.5 (17.70\%) \\
    DSPs & 151 (12.10\%) \\
    CARRY8 & 274 (1.87\%) \\
    \hline
    \end{tabular}
\end{table}

Tab. \ref{tab:main_module_resources} shows the usage of FPGA resources by the designed architecture. The percentage of the total number of available on-chip resources is also given in brackets.
It should be noted that the implementation of the CM algorithm uses only a small part of the overall logic resources of the \textit{Kria KV260 Vision AI Starter Kit} - less than 4\% for Look-up Tables (LUT) and registers. A relatively higher load was observed for Digital Signal Processing (DSP) resources (12.1\%) and BRAM (17.7\%), which is a direct result of the intensive use of arithmetic operations and parallel data buffering in the architecture. The presented resource utilization also indicates that a larger ROI (e.g. \(128 \times 128\)) could be used, which would result in a roughly fourfold increase in BRAM utilization.




According to the power estimation performed using the Vivado Power tool, the FPGA consumed approximately 0.813 W of on-chip power, including both dynamic and static components. This represents a relatively low power demand, especially considering the efficiency and performance achieved.

\section{Conclusions}
\label{sec:conclusions}

In this paper we present a hardware architecture design implementing the CM algorithm for FPGA computing resources.
Experimental results show that the processing time on the FPGA is roughly \(200\times\) shorter than a CPU implementation and around \(450\times\) shorter than a GPU implementation, confirming the advantages of a hardware-based realization of the algorithm, as FPGA devices exploit spatial parallelism and deep pipelining, enabling concurrent processing of data streams at the hardware level. This is the first hardware architecture that allows for acceleration of calculations in the CM algorithm. The operating frequency achieved by the system at 210 MHz, with low utilization of FPGA resources and low power consumption, confirms the high speed and efficiency of the proposed solution.


Further work may include optimizing the architecture to increase the maximum operating frequency of the chip, reducing resource consumption, and modifying the architecture to support more advanced motion models. Another important area for development could be to increase the maximum ROI size, for instance by utilizing external RAM as an additional accumulation buffer. It may also be beneficial to implement a double-buffering mechanism, allowing continuous event acquisition during the optimization phase. The architecture could also benefit from more advanced optimization methods than simple gradient ascent.

\section*{Acknowledgments}
This work was carried out within the Embedded Vision Systems Group at AGH University of Krakow.

%
%
%
\bibliographystyle{splncs04}
\bibliography{mybibliography}

\begin{thebibliography}{10}
\providecommand{\url}[1]{\texttt{#1}}
\providecommand{\urlprefix}{URL }
\providecommand{\doi}[1]{https://doi.org/#1}

\bibitem{asano2009comparison}
Asano, S., Maruyama, T., Yamaguchi, Y.: Performance comparison of fpga, gpu and
  cpu in image processing. In: 2009 international conference on field
  programmable logic and applications. pp. 126--131. IEEE (2009)

\bibitem{gallego2022survey}
Gallego, G., Delbr{\"u}ck, T., Orchard, G., Bartolozzi, C., Taba, B., Censi,
  A., Leutenegger, S., Davison, A.J., Conradt, J., Daniilidis, K., et~al.:
  Event-based vision: A survey. IEEE transactions on pattern analysis and
  machine intelligence  \textbf{44}(1),  154--180 (2020)

\bibitem{gallego2019focus}
Gallego, G., Gehrig, M., Scaramuzza, D.: Focus is all you need: Loss functions
  for event-based vision. In: Proceedings of the IEEE/CVF Conference on
  Computer Vision and Pattern Recognition. pp. 12280--12289 (2019)

\bibitem{gallego2018unifying}
Gallego, G., Rebecq, H., Scaramuzza, D.: A unifying contrast maximization
  framework for event cameras, with applications to motion, depth, and optical
  flow estimation. In: Proceedings of the IEEE conference on computer vision
  and pattern recognition. pp. 3867--3876 (2018)

\bibitem{gehrig2018asynchronous}
Gehrig, D., Rebecq, H., Gallego, G., Scaramuzza, D.: Asynchronous, photometric
  feature tracking using events and frames. In: Proceedings of the European
  Conference on Computer Vision (ECCV). pp. 750--765 (2018)

\bibitem{guo2024cmax}
Guo, S., Gallego, G.: Cmax-slam: Event-based rotational-motion bundle
  adjustment and slam system using contrast maximization. IEEE Transactions on
  Robotics  \textbf{40},  2442--2461 (2024)

\bibitem{hamann2024motion}
Hamann, F., Wang, Z., Asmanis, I., Chaney, K., Gallego, G., Daniilidis, K.:
  Motion-prior contrast maximization for dense continuous-time motion
  estimation. In: European Conference on Computer Vision. pp. 18--37. Springer
  (2024)

\bibitem{kim2021real}
Kim, H., Kim, H.J.: Real-time rotational motion estimation with contrast
  maximization over globally aligned events. IEEE Robotics and Automation
  Letters  \textbf{6}(3),  6016--6023 (2021)

\bibitem{liu2020globally}
Liu, D., Parra, A., Chin, T.J.: Globally optimal contrast maximisation for
  event-based motion estimation. In: Proceedings of the IEEE/CVF Conference on
  Computer Vision and Pattern Recognition. pp. 6349--6358 (2020)

\bibitem{maqueda2018steering}
Maqueda, A.I., Loquercio, A., Gallego, G., Garc{\'\i}a, N., Scaramuzza, D.:
  Event-based vision meets deep learning on steering prediction for
  self-driving cars. In: Proceedings of the IEEE conference on computer vision
  and pattern recognition. pp. 5419--5427 (2018)

\bibitem{mueggler2017event}
Mueggler, E., Rebecq, H., Gallego, G., Delbruck, T., Scaramuzza, D.: The
  event-camera dataset and simulator: Event-based data for pose estimation,
  visual odometry, and slam. The International journal of robotics research
  \textbf{36}(2),  142--149 (2017)

\bibitem{paredes2023taming}
Paredes-Vall{\'e}s, F., Scheper, K.Y., De~Wagter, C., De~Croon, G.C.: Taming
  contrast maximization for learning sequential, low-latency, event-based
  optical flow. In: Proceedings of the IEEE/CVF international conference on
  computer vision. pp. 9695--9705 (2023)

\bibitem{peng2021globally}
Peng, X., Gao, L., Wang, Y., Kneip, L.: Globally-optimal contrast maximisation
  for event cameras. IEEE Transactions on Pattern Analysis and Machine
  Intelligence  \textbf{44}(7),  3479--3495 (2021)

\bibitem{qasaimeh2019comparing}
Qasaimeh, M., Denolf, K., Lo, J., Vissers, K., Zambreno, J., Jones, P.H.:
  Comparing energy efficiency of cpu, gpu and fpga implementations for vision
  kernels. In: 2019 IEEE international conference on embedded software and
  systems (ICESS). pp.~1--8. IEEE (2019)

\bibitem{shiba2022event}
Shiba, S., Aoki, Y., Gallego, G.: Event collapse in contrast maximization
  frameworks. Sensors  \textbf{22}(14), ~5190 (2022)

\bibitem{shiba2024secrets}
Shiba, S., Klose, Y., Aoki, Y., Gallego, G.: Secrets of event-based optical
  flow, depth and ego-motion estimation by contrast maximization. IEEE
  Transactions on Pattern Analysis and Machine Intelligence  \textbf{46}(12),
  7742--7759 (2024)

\bibitem{stoffregen2019event}
Stoffregen, T., Kleeman, L.: Event cameras, contrast maximization and reward
  functions: An analysis. In: Proceedings of the IEEE/CVF Conference on
  Computer Vision and Pattern Recognition. pp. 12300--12308 (2019)

\bibitem{wang2022visual}
Wang, Y., Yang, J., Peng, X., Wu, P., Gao, L., Huang, K., Chen, J., Kneip, L.:
  Visual odometry with an event camera using continuous ray warping and
  volumetric contrast maximization. Sensors  \textbf{22}(15), ~5687 (2022)

\end{thebibliography}

\end{document}